\documentclass[final,5p,times]{elsarticle}
\usepackage[numbers]{natbib}
\usepackage{amsmath,amssymb,amsfonts}
\usepackage{graphicx}
\usepackage{booktabs}
\usepackage{xcolor}

\journal{Pattern Recognition Letters}

\begin{document}

\begin{frontmatter}


\title{Analytical Modeling and Correction of Distance Error in Homography-Based Ground-Plane Mapping}


\author{Mateusz Szulc}
\ead{MateuszSzulc555@wp.pl}

\author{Marcin Iwanowski\corref{cor1}}
\ead{marcin.iwanowski@pw.edu.pl}
\cortext[cor1]{Corresponding author}

\address{Institute of Control and Industrial Electronics, Faculty of Electrical Engineering, Warsaw University of Technology, ul.~Koszykowa~75, 00-662 Warsaw, Poland}

\begin{abstract}
Accurate distance estimation from monocular cameras is essential for intelligent monitoring systems. In many deployments, image coordinates are mapped to ground positions using planar homographies initialized by manual selection of corresponding regions. Small inaccuracies in this initialization propagate into systematic distance distortions. This paper derives an explicit relationship between homography perturbations and the resulting distance error, showing that the error grows approximately quadratically with the true distance from the camera. Based on this model, two simple correction strategies are evaluated: regression-based estimation of the quadratic error function and direct optimization of the homography via coordinate-based gradient descent. A large-scale simulation study with more than 19 million test samples demonstrates that regression achieves higher peak accuracy when the model is reliably fitted, whereas gradient descent provides greater robustness against poor initial calibration. This suggests that improving geometric calibration may yield greater performance gains than increasing model complexity in many practical systems.
\end{abstract}


\end{frontmatter}
\section{Introduction}
Camera-based distance estimation is a fundamental component of intelligent monitoring systems, supporting diverse applications. In many deployed systems, monocular cameras map object positions from image coordinates to real-world distances on the road plane using planar homographies, valued for their low cost and ease of installation.

In practice, these homographies are frequently initialized through manual selection of corresponding regions between the image and a site map---for example, by fitting a trapezoid to the road surface and mapping it to a rectangular reference. While simple and widely adopted, this procedure introduces systematic geometric inaccuracies that propagate into distance estimates, with even small pixel-level errors causing substantial distortions at range. Despite its practical importance, the range-dependent amplification of homography-induced distance error is rarely expressed in an explicit analytical form. Existing approaches typically rely on empirical observations or general uncertainty propagation frameworks, which do not yield a compact relationship between perturbation magnitude and distance error. To the best of our knowledge, no prior work provides a closed-form analytical derivation that explicitly links homography perturbation to distance estimation error as a function of range under minimal geometric assumptions. Existing approaches either rely on empirical observations or general uncertainty propagation frameworks, which do not yield a compact and directly interpretable relationship.

This paper focuses on trapezoid selection error as a distinct source of geometric uncertainty, assuming that intrinsic calibration and lens distortion correction have already been performed and that the scene can be approximated by a dominant planar road surface. We derive an explicit analytical relationship between manual homography initialization errors and the resulting distance estimation error, showing that errors grow approximately quadratically with range---a behavior commonly observed in practice but rarely formalized. Importantly, this result reveals that the dominant source of distance estimation error is not algorithmic but geometric, arising directly from the projective structure of the mapping.

Based on this analysis, we evaluate two practical correction strategies. The first models the distance error as a quadratic function and estimates its parameters via regression from three calibration points. The second directly optimizes the trapezoid coordinates using coordinate-based gradient descent to minimize calibration error. Both approaches operate on top of standard homography-based pipelines without requiring additional sensors or changes to detection and tracking stages.

We conduct a large-scale evaluation in a controlled simulation environment using an ideal pinhole camera model, with more than 19 million test samples across two simulated traffic scenes.

The main contributions of this work are:

\begin{itemize}
\item a closed-form analytical model linking homography perturbation to distance estimation error,
\item a derivation showing that the dominant error component follows a quadratic law with respect to range,
\item practical correction strategies demonstrating how the analytical model can be exploited in real systems.
\end{itemize}

It reveals a previously unformalized mechanism of error amplification inherent to planar homography mappings, which has direct implications for the reliability of monocular distance estimation in real-world deployments. This insight enables lightweight post-hoc correction mechanisms and suggests that improving geometric calibration may yield greater gains than increasing model complexity in many practical systems.

\section{Related Work}

The homography-based approach described above is widely used in practice, with object detections reduced to representative ground-contact points and projected into a metric coordinate frame for further analysis and tracking.

Ground-plane localization has been extensively studied in multi-camera traffic and surveillance systems, where homography-based mappings fuse detections from different viewpoints in a common reference frame \cite{KhanShahECCV2006,KimDavisECCV2006,KhanShahPAMI2009,SantosMorimotoPRL2011,FleuretPAMI2008}, supporting robust tracking and trajectory estimation under occlusions \cite{BlackEllisIVC2006,MunozSalinasPatternRec2010,YinIETCV2015}. Some approaches instead transform the entire image into a bird's-eye-view via inverse perspective mapping (IPM) and perform detection or tracking in this domain \cite{BertozziBroggiFascioliIVC1998,OliveiraInfFus2015}, though these remain fundamentally dependent on accurate planar mappings \cite{BrulsArxiv2018,TanveerSgorbissaICPRAM2018}.

A substantial body of work addresses automatic and weakly supervised camera calibration for traffic scenes, exploiting scene geometry, vanishing points, or object height priors to recover projection parameters without dedicated targets \cite{LvZhaoNevatiaPAMI2006,MicusikPajdlaCVPR2010,DubskaBMVC2014,AutoCalibBuildSys2017,CCTVCalibMVAP2023}. Even so, practical implementations often rely on manual initialization of planar correspondences.

The impact of geometric uncertainty on planar localization accuracy has been studied from a probabilistic perspective. Homography estimation errors have been analyzed in terms of reprojection sensitivity and point perturbation \cite{ChumCVIU2005}, and tracking systems explicitly propagate image-space uncertainty into ground-plane covariance models \cite{BlackEllisIVC2006,MunozSalinasPatternRec2010,ClaasenDeVilliersESWA2025}. In this framework, a representative image point $\mathbf{u}=[u,v,1]^T$ is mapped to ground-plane coordinates $\tilde{\mathbf{x}}=\mathbf{H}\mathbf{u}$ and dehomogenized to $\mathbf{x}=[x,y]^T$. The planar covariance is approximated by

\begin{equation}
\mathbf{\Sigma}_{\mathbf{x}} \approx \mathbf{J}_{\mathbf{z}}\,\mathbf{\Sigma}_{\mathbf{z}}\,\mathbf{J}_{\mathbf{z}}^{T}
\;+\;
\mathbf{J}_{\mathbf{h}}\,\mathbf{\Sigma}_{\mathbf{h}}\,\mathbf{J}_{\mathbf{h}}^{T},
\label{eq:cov_propagation}
\end{equation}

where $\mathbf{\Sigma}_{\mathbf{z}}$ models detection noise, $\mathbf{\Sigma}_{\mathbf{h}}$ models homography parameter uncertainty, and $\mathbf{J}_{\mathbf{z}}$, $\mathbf{J}_{\mathbf{h}}$ are the corresponding Jacobians. This decomposition is commonly used for Kalman filtering on the ground plane, Mahalanobis-distance gating, and quantifying calibration drift \cite{ClaasenDeVilliersESWA2025,BlackEllisIVC2006,MunozSalinasPatternRec2010}.

Despite these advances, the covariance propagation framework in~\eqref{eq:cov_propagation} captures the general structure of this uncertainty but does not yield a closed-form expression relating trapezoid perturbation magnitude to the resulting distance error at a given range. The following section derives this relationship explicitly.
This suggests that existing formulations are well-suited for uncertainty quantification but not for explaining systematic geometric error growth. In contrast to these approaches, the present work focuses on isolating and explicitly modeling a deterministic geometric error component, providing a compact analytical relationship that complements existing probabilistic formulations.

\section{Theoretical Derivation of Quadratic Error}

Building on the homography formulation reviewed above, this section derives an explicit relationship between trapezoid perturbation and the resulting distance estimation error, providing the analytical foundation for the correction strategies presented in Section~4.

\begin{figure}[t]
\centerline{\includegraphics[width=0.7\columnwidth]{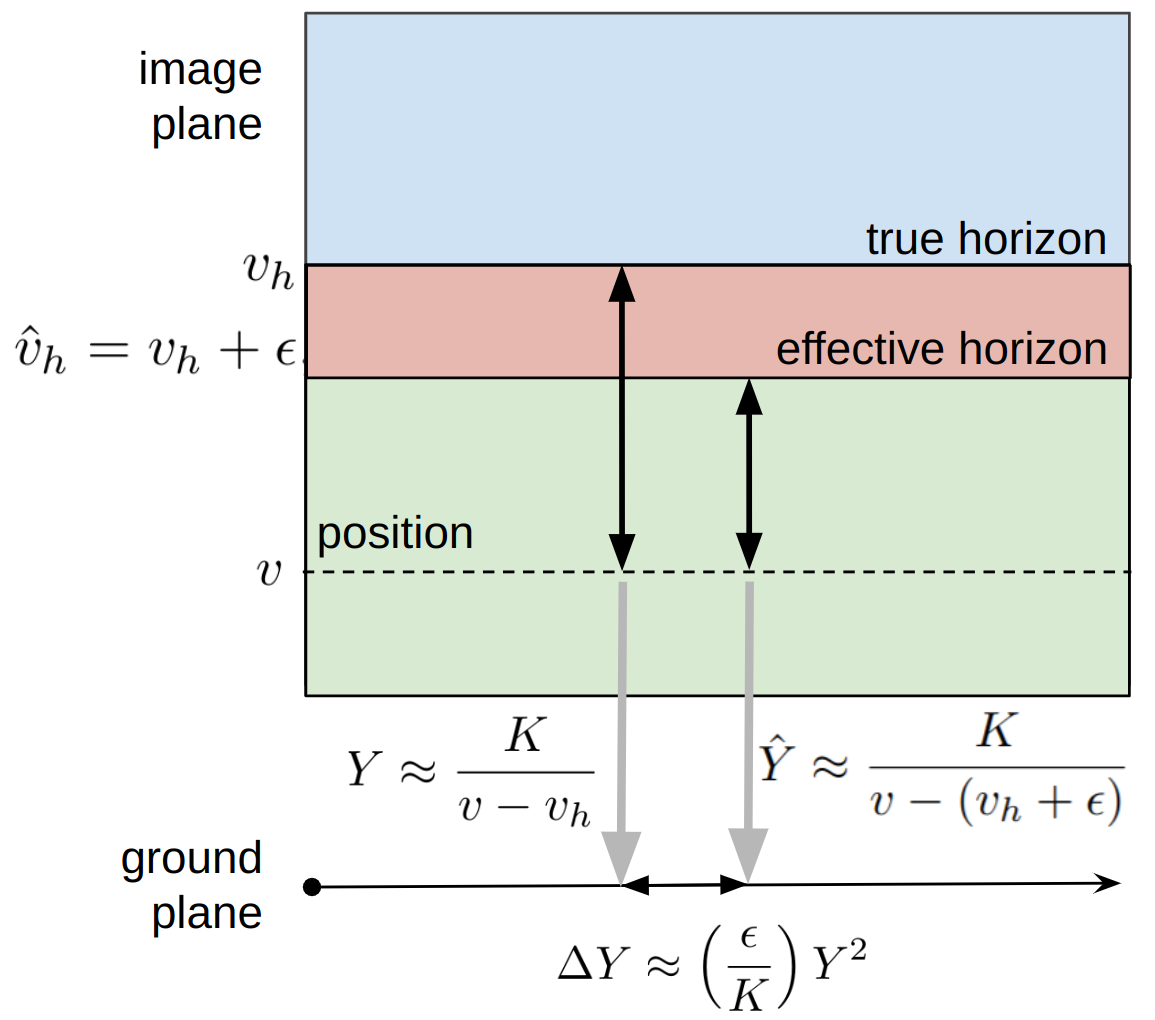}}
\caption{Minimal geometry of homography-based depth mapping. A small perturbation of the effective horizon position changes the denominator $(v - v_h)$ in the projective depth model, which leads to increasing distance error at larger ranges.}
\label{fig:geometry}
\end{figure}

While prior work models uncertainty propagation through homography estimation using covariance formulations, these approaches remain general and do not provide an explicit range-dependent error law. Fig.~\ref{fig:geometry} illustrates the minimal geometric configuration underlying the proposed analysis, highlighting how a perturbation of the effective horizon position leads to increasing distance error at larger ranges.

The perspective mapping from pixel coordinates $(u, v)$ to world coordinates $(X, Y)$ is expressed in terms of a homography matrix. Since in our case we focus on the estimation of distance from the camera, only the $Y$ coordinate is meaningful, where coefficients $h_{ij}$ are defined as:
\begin{equation}
Y(u,v) = \frac{h_{21}u + h_{22}v + h_{23}}{h_{31}u + h_{32}v + h_{33}}
\end{equation}
Since each image column corresponds to a fixed world column $X = \text{const}$, we are only interested in how $Y$ varies with the row coordinate $v$ for a given $u$. All terms involving $u$ therefore collapse into constant coefficients, and the expression for $Y$ simplifies to:
\begin{equation}
Y(v) = \frac{A v + B}{C v + D},
\end{equation}
where $A = h_{22}$, $B = h_{21}u + h_{23}$, $C = h_{32}$, and $D = h_{31}u + h_{33}$ are constants for a given column $u$.
The horizon position $v_h$ is defined by the zero of the denominator:
\begin{equation}
C v_h + D = 0 \;\;\Rightarrow\;\; D = -C v_h.
\end{equation}
Substituting this expression for $D$ yields:
\begin{equation}
Y(v) = \frac{A v + B}{C (v - v_h)}.
\end{equation}
To separate the constant and hyperbolic terms, we add and subtract $A v_h$ in the numerator to facilitate division:
\begin{equation}
Y(v) = \frac{A(v - v_h) + (A v_h + B)}{C (v - v_h)} = \frac{A}{C} + \frac{(A v_h + B)/C}{v - v_h}.
\end{equation}
By defining the constant $\text{Offset} = A/C$ and the scale factor $K = (A v_h + B)/C$, the relationship becomes:
\begin{equation}
Y(v) = \text{Offset} + \frac{K}{v - v_h}.
\end{equation}
Since the constant offset cancels in error analysis or represents a fixed translation, the fundamental relationship simplifies to:
\begin{equation}
Y \approx \frac{K}{v - v_h}.
\label{eq:fundamental}
\end{equation}


An error in trapezoid selection produces a perturbed homography that shifts the estimated horizon by $\epsilon$, such that
\begin{equation}
\hat{v}_h = v_h + \epsilon.
\end{equation}
The resulting estimated distance is
\begin{equation}
\hat{Y} = \frac{K}{v - (v_h + \epsilon)}.
\end{equation}


The distance estimation error is defined as the difference between the estimated distance $\hat{Y}$ and the true distance $Y$:
\begin{equation}
\Delta Y = \hat{Y} - Y = K\!\left(\frac{1}{(v - v_h) - \epsilon} - \frac{1}{v - v_h}\right).
\end{equation}
For a small pixel error relative to the distance from the horizon ($\epsilon \ll v - v_h$), we apply a first-order Taylor expansion with $q = v - v_h$:
\begin{equation}
(q - \epsilon)^{-1} = \frac{1}{q} + \frac{\epsilon}{q^2} + \mathcal{O},
\end{equation}
where $\mathcal{O}$ represents higher-order terms. Since $\epsilon$ is typically on the order of a single pixel and $q$ is much larger, these terms are negligible and are thus omitted. This yields:
\begin{equation}
\Delta Y \approx \frac{K \epsilon}{(v - v_h)^2}.
\label{eq:error_v}
\end{equation}
From~\eqref{eq:fundamental}, we have the relationship $(v - v_h) = K/Y$, which implies:
\begin{equation}
\frac{1}{(v - v_h)^2} = \frac{Y^2}{K^2}.
\label{eq:pixel_to_world}
\end{equation}
Substituting~\eqref{eq:pixel_to_world} into~\eqref{eq:error_v} gives the final result:
\begin{equation}
\boxed{\Delta Y \approx \left(\frac{\epsilon}{K}\right) Y^2.}
\label{eq:quadratic_result}
\end{equation}
Thus, the distance estimation error grows approximately quadratically with the true distance $Y$. The coefficient $\epsilon/K$ reflects the magnitude of homography perturbation normalized by the scale factor. This relationship provides the theoretical motivation for the correction strategies evaluated in this work.
\section{Proposed Correction Methodologies}
Two complementary correction strategies are evaluated, both relying on three reference measurements --- the minimum needed to fit the quadratic error structure derived in Section~3.

\subsection{Method~1: Error Function Approximation (Regression)}

This approach estimates the quadratic error function directly from the calibration samples and applies a spatially directed correction to all subsequent distance estimates.


Let $\mathbf{p}_i = (x_i, y_i)$ denote the position of the $i$-th calibration object after homography projection and scaling, and let $\mathbf{g}_i = (g_{x,i}, g_{y,i})$ denote the corresponding ground-truth position. The Euclidean error at each calibration point is

\begin{equation}
e_i = \|\mathbf{p}_i - \mathbf{g}_i\| = \sqrt{(x_i - g_{x,i})^2 + (y_i - g_{y,i})^2}.
\label{eq:euclidean_error}
\end{equation}

For each calibration point, the distance from the camera origin $\mathbf{o} = (o_x, o_y)$ on the ground plane is computed as $d_i = \|\mathbf{p}_i - \mathbf{o}\|$. The camera origin $\mathbf{o}$ corresponds to the projection of the camera's optical center onto the ground plane, which in a standard homography-based pipeline is known from the mounting geometry or can be inferred from the vanishing point of vertical lines.


Based on the theoretical result~\eqref{eq:quadratic_result}, the relationship between range and error magnitude is modeled as
\begin{equation}
e(d) = a\,d^2 + b\,d,
\label{eq:quadratic_model}
\end{equation}
where $a$ and $b$ are scalar coefficients to be estimated. The intercept term is omitted because the error must vanish at the camera origin ($d = 0$). Given three calibration pairs $(d_i, e_i)$, the parameters are obtained by solving the overdetermined linear system
\begin{equation}
\begin{bmatrix}
d_1^2 & d_1 \\
d_2^2 & d_2 \\
d_3^2 & d_3
\end{bmatrix}
\begin{bmatrix} a \\ b \end{bmatrix}
=
\begin{bmatrix} e_1 \\ e_2 \\ e_3 \end{bmatrix}
\label{eq:normal_equations}
\end{equation}

in the least-squares sense via the normal equations. Specifically, defining

\begin{equation}
S_{k} = \sum_{i=1}^{3} d_i^k, \quad T_k = \sum_{i=1}^{3} d_i^k\, e_i,
\end{equation}

the solution is

\begin{equation}
a = \frac{T_2\, S_2 - T_1\, S_3}{S_4\, S_2 - S_3^2}, \qquad
b = \frac{S_4\, T_1 - S_3\, T_2}{S_4\, S_2 - S_3^2},
\label{eq:closed_form_ab}
\end{equation}

\noindent provided the determinant $S_4\, S_2 - S_3^2 \neq 0$, which holds whenever the three calibration distances are distinct. Non-negativity constraints $a \geq 0$, $b \geq 0$ are enforced by clamping, since negative coefficients would imply correction in the wrong direction. If the system is degenerate (e.g., all calibration points at nearly the same distance), the method falls back to a linear-only model $e(d) = b\,d$.

With only three equations in two unknowns, the system in~\eqref{eq:normal_equations} is only mildly overdetermined, so the quadratic fit is tightly constrained by the available data. This makes the method highly efficient but also sensitive to the placement of calibration points: if all three points lie at similar distances, the quadratic component $a$ is poorly determined, while widely separated points improve curvature estimation at the cost of potential noise amplification at the extremes.


The scalar correction $c_i = a\,d_i^2 + b\,d_i$ is applied along the unit direction vector from each mapped point toward the camera origin:

\begin{equation}
\hat{\mathbf{d}}_i = \frac{\mathbf{o} - \mathbf{p}_i}{\|\mathbf{o} - \mathbf{p}_i\|}.
\label{eq:direction_vector}
\end{equation}

The corrected position is then

\begin{equation}
\mathbf{p}_i^{\,\text{corr}} = \mathbf{p}_i - \hat{\mathbf{d}}_i \cdot c_i.
\label{eq:corrected_position}
\end{equation}

This directional formulation extends the one-dimensional theoretical result from Section~3 to the two-dimensional ground plane. The correction displaces each point radially toward or away from the camera by an amount proportional to the estimated error at its range, preserving the angular position of the object relative to the camera. This is consistent with the physical mechanism of the error: homography perturbations primarily affect depth (range) estimation while leaving lateral position relatively unaffected.

Once $a$ and $b$ have been estimated from the three calibration points, the correction in~\eqref{eq:corrected_position} is applied to every mapped detection in the scene without further parameter updates.

\subsection{Method~2: Trapezoid Coordinate Optimization (Gradient Descent)}

The second strategy directly adjusts the trapezoid vertex coordinates that define the homography, rather than modeling the output error. The four vertices of the video-side trapezoid are treated as optimization variables and perturbed to minimize calibration error on the same three reference points.


The objective function is the mean absolute distance error evaluated on the selected calibration subset:

\begin{equation}
\mathcal{L}(\mathbf{v}) = \frac{1}{|\mathcal{S}|}\sum_{i \in \mathcal{S}} \left| e_i(\mathbf{v}) \right|,
\label{eq:gd_objective}
\end{equation}

where $\mathbf{v} \in \mathbb{R}^{8}$ collects the $(x,y)$ coordinates of the four trapezoid vertices, $\mathcal{S}$ is the set of calibration point indices, and $e_i(\mathbf{v})$ is the Euclidean error~\eqref{eq:euclidean_error} evaluated using the homography induced by~$\mathbf{v}$. Each candidate vertex configuration is constrained to lie within a maximum displacement $\delta_{\max}$ of the initial position to prevent degenerate homographies.


The optimization proceeds in two phases of decreasing granularity:
\textit{Phase~1 (Coarse search).} A subset of geometrically influential vertices (specifically, two opposing corners of the trapezoid) is perturbed over a coarse grid of candidate displacements. This phase identifies a promising basin of attraction at low computational cost while focusing on the vertices that most strongly affect the perspective mapping.
\textit{Phase~2 (Fine coordinate descent).} All four vertices are optimized using coordinate descent with progressively decreasing step sizes. At each step, a single coordinate of a single vertex is perturbed by $\pm\delta$, and the change is accepted if it reduces $\mathcal{L}$. The update order is randomized at each pass to avoid directional bias. The step size is reduced when no improvement is found in consecutive passes, and optimization terminates when the smallest step size is exhausted or a computational budget is reached.

\section{Simulation Setup and Evaluation Framework}

To isolate geometric effects from confounding factors such as illumination variation and sensor uncertainty, all experiments were conducted in a controlled simulation environment based on Unreal Engine~5.4. Simulation-based evaluation is widely adopted in vision and robotics research to enable precise ground-truth comparison under repeatable conditions. In this setup, object positions recorded in the simulator serve as exact references for evaluating distance estimation accuracy, eliminating disturbances commonly present in real-world video data. The magnitude of simulated perturbations (0–3 pixels) was selected to reflect typical inaccuracies observed in manual homography initialization in real-world deployments, ensuring that the evaluated error regime remains practically relevant.

\subsection{Camera and Detection Parameters}
Scenes were captured using an ideal pinhole camera with the following parameters:  \textbf{Field of view:} $95^\circ$
, \textbf{Mounting height:} $4\,\mathrm{m}$, \textbf{Tilt angle:} $15^\circ$ downward, \textbf{Lens distortion:} none -- distortion correction is assumed to be performed in practical deployments, and the present study focuses exclusively on planar mapping error, \textbf{Terrain:} perfectly planar, in order to isolate homography-induced error from elevation effects.

Object detections were obtained using YOLOv6l at 20~fps and matched to the temporally nearest ground-truth positions recorded in the simulation at 50~ms intervals. The use of an ideal camera model is intentional, as the objective of this study is to analyze and correct errors arising from homography initialization rather than from intrinsic calibration or lens distortion.
\subsection{Testing Framework and Error Generation}

For each scene, an initial trapezoid was defined manually to approximate the road surface. Despite careful placement, these initial trapezoids contained residual geometric errors. To evaluate robustness with respect to calibration inaccuracy, 1{,}000 additional trapezoid variants were generated per scene. In each variant, a random perturbation of 0, 1, 2, or 3 pixels was applied to a randomly selected subset of 2 to 16 coordinate values (across two trapezoids with four vertices each, where each vertex is represented by $(x,y)$ coordinates). Both correction methods were evaluated using identical perturbations to ensure fair comparison.

A blind evaluation protocol was adopted: only three calibration samples were used to estimate correction parameters, and no information from the remaining trajectories was available during the correction stage. Calibration samples were placed directly in front of the camera at distances ranging from 5~m to 44~m in Scene~1 and from 5~m to 49~m in Scene~2. This setup reflects practical deployment conditions in which only sparse calibration data are available or large camera networks must be calibrated under time constraints.

\begin{figure}[htbp]
\centerline{\includegraphics[width=0.6\columnwidth]{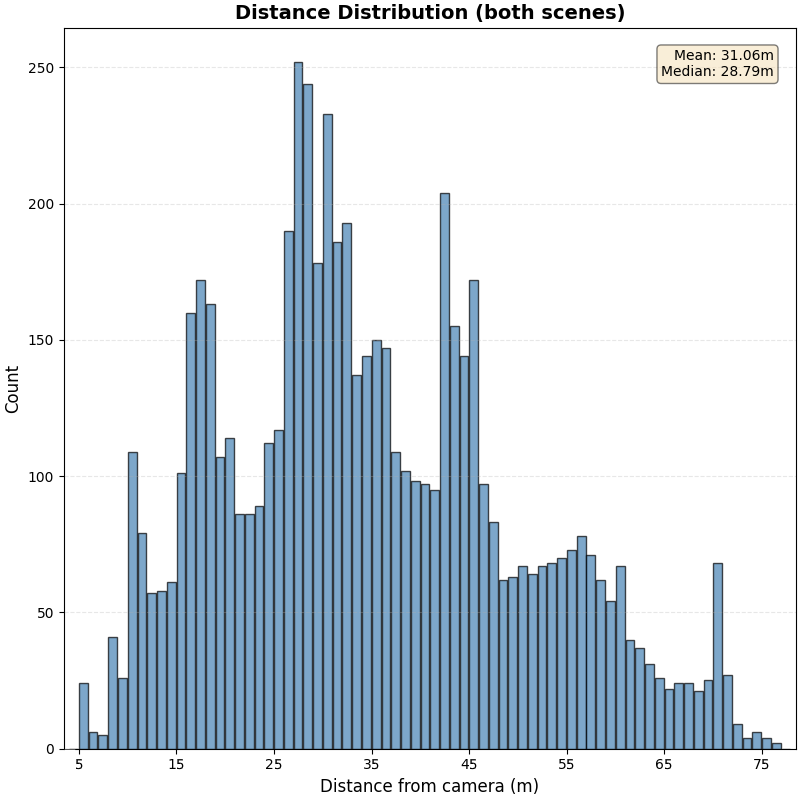}}
\caption{Distribution of object distances from the camera in simulated pedestrian traffic, defining the operating range used for evaluation.}
\label{fig:traffic_distance_distribution}
\end{figure}

The test trajectories follow a distance distribution shown in Fig.~\ref{fig:traffic_distance_distribution}. This distribution defines the effective operating range of the system and serves as a representative test set for evaluating generalization beyond the calibration points.

\begin{figure}[htbp]
\centerline{\includegraphics[width=0.6\columnwidth]{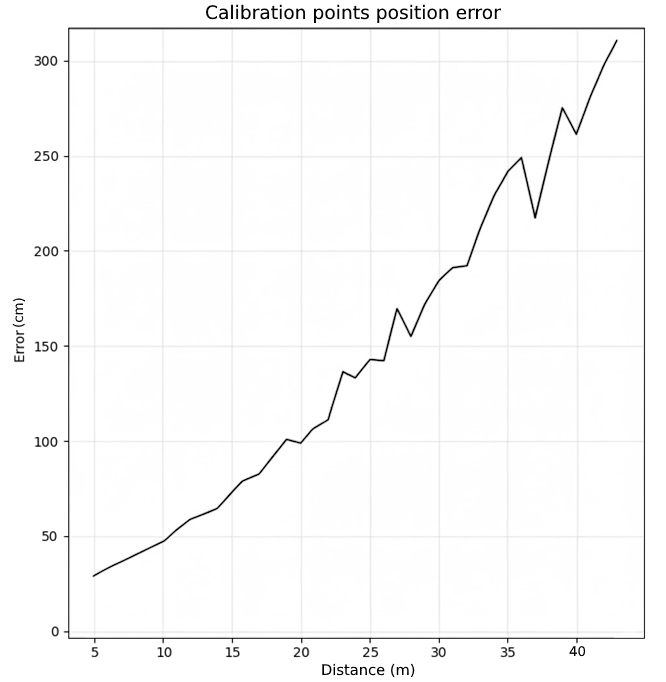}}
\caption{Example of distance-dependent calibration error, illustrating inconsistencies that may lead to overfitting on calibration points.}
\label{fig:calibration_error_diagram}
\end{figure}

Calibration error is not uniformly distributed across the distance range, and localized anomalies may occur that allow a correction model to fit the calibration points while failing to generalize to full trajectories (Fig.~\ref{fig:calibration_error_diagram}). Characterizing this risk is central to evaluating the robustness of the proposed correction strategies.

This example highlights that consistency on calibration points does not necessarily translate into trajectory-level accuracy, emphasizing the importance of evaluating correction methods beyond sparse calibration samples.

\section{Experimental Results}

The results confirm that both methods substantially reduce distance estimation error, but exhibit a clear trade-off between peak accuracy and robustness.

Both correction strategies were evaluated on 19{,}038{,}019 samples generated from 1{,}000 trapezoid perturbations and all combinations of three calibration points. Performance was quantified as relative improvement in distance estimation error on independent test trajectories; aggregate results are summarized in Table~\ref{tab:method_comparison}.

The regression-based method yielded positive path improvement in 93.0\% of cases and positive calibration improvement in 96.5\% of cases. The mean path improvement was 73.9\%, with a median of 90.0\%, indicating strong performance under typical conditions but the presence of a non-negligible tail of failure cases. However, the improvement distributions reveal a notable asymmetry: calibration improvement is concentrated in the upper range (median 95.4\%, interquartile range~20.0 percentage points), whereas path improvement is substantially more dispersed (median 89.6\%, standard deviation~47.4 percentage points), with a heavy negative tail extending below $-100\%$ in worst-case configurations. This gap reflects the fundamental risk of overfitting: the quadratic model can fit the three calibration points well while mischaracterizing the error function over the full operating range.

The gradient descent method achieved positive calibration improvement in 100\% of cases by construction and positive path improvement in 98.8\% of cases. The mean path improvement was 76.5\%, with a median of 86.1\%. Compared with regression, gradient descent exhibited greater robustness against catastrophic degradation, at the cost of slightly lower median performance in favorable scenarios. The correlation between calibration and path improvement was lower (0.56), indicating reduced predictability of trajectory-level improvement from calibration data alone. The calibration improvement distribution for gradient descent is tightly concentrated (median 90.9\%, standard deviation~8.6 percentage points), and the path improvement distribution, while broader (standard deviation~23.8 percentage points), exhibits a far less severe negative tail than regression, with the first percentile at $-3.8\%$ compared to $-135\%$ for regression.

Both methods exhibited a systematic dependence between calibration improvement and trajectory-level performance. When calibration improvement exceeded approximately 75\%, regression achieved higher average gains, while gradient descent provided consistently positive results across all tested cases.

\begin{figure}[htbp]
\begin{tabular}{cc}
 \includegraphics[width=0.49\columnwidth]{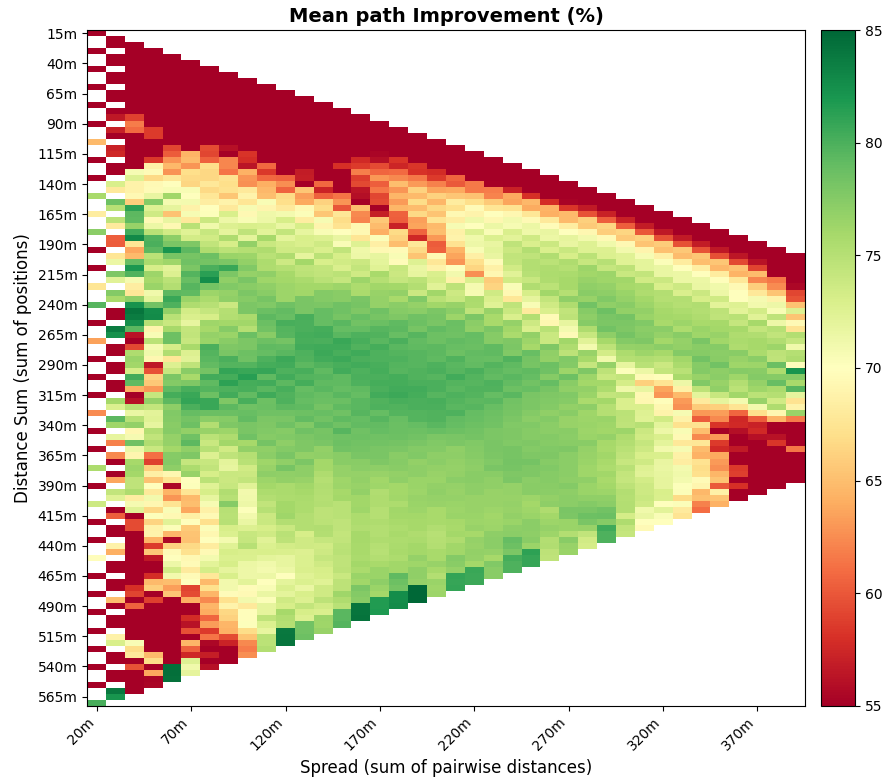}    &  \includegraphics[width=0.49\columnwidth]{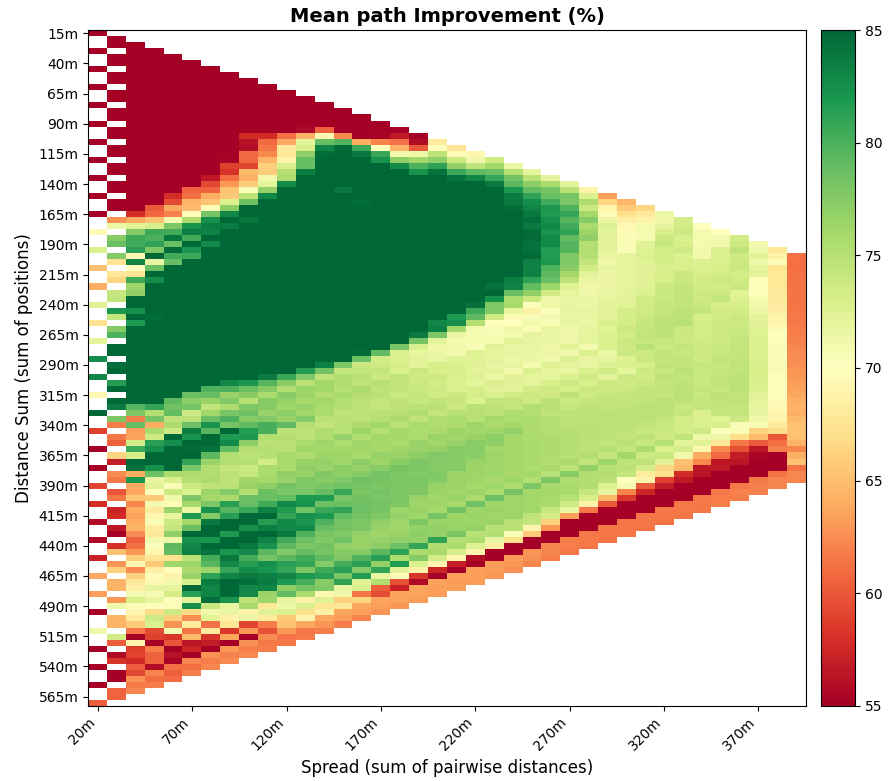}\\
  (a)   &  (b)
\end{tabular}
\caption{Mean path improvement as a function of total distance sum and spread for the regression-based (a) and gradient-descent (b) methods.}
\label{fig:heatmap_improvement}
\end{figure}

Calibration point selection strongly influenced both methods (Fig.~\ref{fig:heatmap_improvement}). Points placed too close to the camera produced insufficient error signal, whereas points at large distances amplified noise. Likewise, insufficient spatial spread limited the ability to capture the quadratic structure of the error function. Gradient descent showed greater sensitivity to unfavorable point placement than regression, reflecting its reliance on only three calibration points to guide trapezoid optimization.

\begin{table}[t]
\centering
\footnotesize
\caption{Summary comparison of correction methods.}
\label{tab:method_comparison}
\begin{tabular}{lcc}
\toprule
Metric & Regression & Grad.\ Descent \\
\midrule
Positive path improvement    & 93.0\%  & 98.8\%  \\
Positive calib.\ improvement & 96.5\%  & 100.0\% \\
Mean path improvement        & 73.9\%  & 76.5\%  \\
Median path improvement      & 90.0\%  & 86.1\%  \\
Path--calib.\ correlation    & 0.84    & 0.56    \\
Failure risk                 & higher  & lower   \\
\bottomrule
\end{tabular}
\end{table}

Figure~\ref{fig:heatmap_improvement}(a) illustrates the mean path improvement obtained with regression-based correction as a function of total distance sum and spread of the calibration points.  A clear optimal calibration regime is visible at moderate distances and sufficient spread.

Figure~\ref{fig:heatmap_improvement}(b) shows the corresponding results for the gradient descent method. Although a similar optimal region is present, performance degrades more rapidly for extreme distances and low spread, indicating increased susceptibility to overfitting on the calibration points.

\section{Discussion}

The experimental findings confirm the approximately quadratic growth of distance estimation error with range, consistent with the analytical model in Section~3. Importantly, this behavior emerges directly from the projective geometry of the mapping and is therefore independent of the specific detection, tracking, or learning components used in the system. This indicates that a substantial portion of error in practical systems may be attributed to geometric miscalibration rather than limitations of detection or learning-based components.


When the regression fit is reliable, it achieves higher peak accuracy and higher median trajectory improvement than gradient descent. However, an inaccurate quadratic coefficient leads to systematic over- or under-compensation at range, producing the heavy failure tail observed in the path improvement distribution. The gradient descent strategy guarantees non-negative improvement on calibration samples by construction, explaining its high proportion of positive trajectory outcomes and robustness against catastrophic failures. The weaker calibration--path correlation (0.56 vs.\ 0.84) indicates, however, that minimizing error on a sparse set of points does not ensure global optimality, and local trapezoid adjustments may overfit poorly distributed calibration points.

The point selection analysis provides a geometric interpretation of these effects. Points too close to the camera yield small absolute errors that under-determine the quadratic component; distant points are dominated by noise and exert disproportionate influence. Limited spatial spread prevents reliable separation of linear and quadratic contributions. These mechanisms explain the ``optimal calibration zone'' visible in the heatmaps and indicate that calibration strategies should be adapted to the dominant operating range of the monitored road segment.

From an operational standpoint, these complementary characteristics motivate a simple hybrid strategy: run both methods, evaluate regression's calibration improvement, and use regression if it exceeds approximately 75\% --- otherwise fall back to gradient descent, which offers lower peak accuracy but avoids catastrophic degradation.

Several limitations should be acknowledged. The analysis assumes a single dominant road plane and neglects intrinsic calibration and lens distortion uncertainty. The controlled simulation environment does not capture all complexities of real traffic scenes, and the gradient descent configuration may not be exhaustively optimized.

Importantly, the proposed analytical relationship isolates a purely geometric effect, independent of detection noise, learning-based components, or scene variability, which both enables controlled analysis and highlights the fundamental nature of the observed error mechanism.

\section{Conclusion}

This paper provides a closed-form analytical characterization of distance estimation error in homography-based ground-plane mapping, showing that the dominant component grows approximately quadratically with range, and evaluated two correction strategies exploiting this structure using only three reference measurements. Regression achieves higher peak accuracy when the quadratic fit is reliable; gradient descent offers greater robustness and a far less severe failure tail. Their complementary strengths support a hybrid strategy: prefer regression when calibration improvement is high, otherwise fall back to gradient descent.

Future work includes incorporating intrinsic calibration uncertainty and lens distortion, validating on real-world traffic datasets, developing adaptive calibration point selection algorithms, investigating online calibration updates, and enforcing multi-camera consistency constraints. The proposed formulation is lightweight, interpretable, and directly applicable in practical systems, offering a simple alternative to more complex calibration refinement procedures. The framework is compatible with standard homography-based pipelines and requires no additional sensors or modifications to detection and tracking components. This finding highlights that improving geometric calibration may be more impactful than increasing model complexity in many practical systems.

\bibliographystyle{elsarticle-num}

\end{document}